\documentclass{article}
\usepackage{spconf,amsmath,graphicx}

\usepackage{array}
\usepackage{multirow}
\usepackage{amsmath}
\usepackage{amsfonts}
\usepackage{graphicx}
\usepackage{xcolor}
\usepackage{color}
\usepackage{enumerate}
\usepackage{arydshln}
\usepackage{booktabs}
\usepackage{tikz}
\usepackage{algorithm}
\usepackage{subfig}
\usepackage{multirow}
\usepackage[noend]{algorithmic}
\usepackage{tablefootnote}
\usepackage{hyperref}

\title{UniSpeech-SAT: Universal Speech Representation Learning with  Speaker Aware Pre-Training}
\name{\begin{tabular}{c}Sanyuan Chen$^{1,2}$\thanks{Work done during an internship at Microsoft.},  Yu Wu$^2$, Chengyi Wang$^2$, Zhengyang Chen$^2$, Zhuo Chen$^2$, Shujie Liu$^2$, \\ Jian Wu$^2$, Yao Qian$^2$, Furu Wei$^2$, Jinyu Li$^2$, Xiangzhan Yu$^{1}$ \end{tabular} }

\address{$^1$Harbin Institute of Technology, China, $^2$Microsoft Corporation}
\begin{document}
%
\maketitle
\begin{abstract}
Self-supervised learning (SSL) is a long-standing goal for speech processing, since it utilizes large-scale unlabeled data and avoids extensive human labeling. Recent years witness great successes in applying self-supervised learning in speech recognition, while
limited exploration was attempted in applying SSL for modeling speaker characteristics. 
In this paper, we aim to improve the existing SSL framework for speaker representation learning. Two methods are introduced for enhancing the unsupervised speaker information extraction. First, we apply the multi-task learning to the current SSL framework, where we integrate the utterance-wise contrastive loss with the SSL objective function.
Second, for better speaker discrimination, we propose an utterance mixing strategy for data augmentation, where additional overlapped utterances are created unsupervisely and incorporate during training. 
We integrate the proposed methods into the HuBERT framework. Experiment results on SUPERB benchmark show that the proposed system achieves state-of-the-art performance in universal representation learning, especially for speaker identification oriented tasks. An ablation study is performed verifying the efficacy of each proposed method. 
Finally, we scale up training dataset to 94 thousand hours public audio data and achieve further performance improvement in all SUPERB tasks.

\end{abstract}

\section{Introduction}
Self-supervised learning has achieved great successes in natural language processing, which utilizes a large amount of unlabeled data to learn universal representation.  The representation enjoys outstanding  generalizability, re-usability, and effectiveness, thus brings significant performance improvements when employed by various downstream tasks. Motivated by this, a series of work in speech processing have been proposed to leverage unlabeled audio for representation learning.  

Self-supervised learning methods are categorized into discriminative methods \cite{cpc, modified_cpc, wav2vec, vq_wav2vec, wav2vec2,hsu2021hubert,zhang2021bigssl}, generative methods \cite{apc1,apc2,vq_apc,npc,mockingjay,tera}, and multi-task learning methods \cite{pase+}. The typical generative method is Autoregressive Predictive Coding (APC) \cite{apc1,apc2}, where the model is similar to the autoencoder
architectures except that the network is trained to predict features for future time steps. The discriminative methods usually employ contrastive learning \cite{wav2vec2} or classification on weak clustering label \cite{hsu2021hubert} to pre-train an encoder network with large-scale unsupervised data.
Recently, the discriminative methods achieved great successes in automatic speech recognition (ASR), which outperforms the best system for Librispeech dataset in 2019 with significantly less supervised data.  The improved performance on different speech tasks in SUPERB benchmark \cite{superb}
also verifies the effectiveness of pre-training. 

Although achieving numerous successes, most pretraining methods for speech application focus on the extraction of spoken content information, i.e. learning representation optimized for tasks such as speech recognition, keyword spotting, etc. 
Limited exploration was carried out on other speech characteristics. 
As speech signal contains multi-fold information, e.g. content, identity, presentation etc., optimization for one aspect might lead to sub-optimized representation for other tasks.
Interestingly, even trained with ASR-oriented objective function, the representation learnt by unsupervised pre-training shows excellent performance in speaker identification related tasks, such as speaker verification, diarization etc., in SUPERB challenge. 
However, can the speaker tasks' performance be further boosted, when provided embedding from matched pre-training, is still an open question.

To answer this question,  we investigate the unsupervised speaker pre-training methods that encourage the preservation of speaker identity. 
Specifically, we proposed two training methods: 
1) We integrate the utterance-wise contrastive loss with the unsupervised representation learning, where the aggregated embedding from each utterance is employed for affinity computation, and a speaker-wise pseudo label is applied as reference.
2) We propose an utterance-mixing training strategy, where partially overlapped signal is constructed for each training sample, by mixing it with a randomly selected speech piece, while the training objective remains the same. 
We integrate our proposed training method in the HuBERT framework \cite{hsu2021hubert}, and conduct experiment on Speech processing Universal PERformance Benchmark (SUPERB) \cite{superb}. The experiment results show that our method significantly improves speaker identification, speaker verification, speaker diarization, and emotion recognition, while maintaining the same speech recognition performance. Finally, we extend our pre-training network to 94k hours of public English audio data, consisting of LibriVox \cite{librilight}, GigaSpeech \cite{GigaSpeech2021}, and VoxPopuli \cite{wang2021voxpopuli}, which further increases performance on speaker tasks compared to previous work using 60k LibriVox data only.

The contribution of the paper is summarized into three-folds. 1) We propose a speaker aware pre-training method which is complementary to current ASR oriented pre-training. 2) We empirically evaluate the model performance on the SUPERB benchmark and achieve state-of-the-art performance in the overall evaluation. 3) We release our model at 
\url{https://github.com/microsoft/UniSpeech}.

\begin{figure*}[t]
	\centering
	\begin{tikzpicture}
	\draw (0,0 ) node[inner sep=0] {\includegraphics[width=11.5cm, clip]{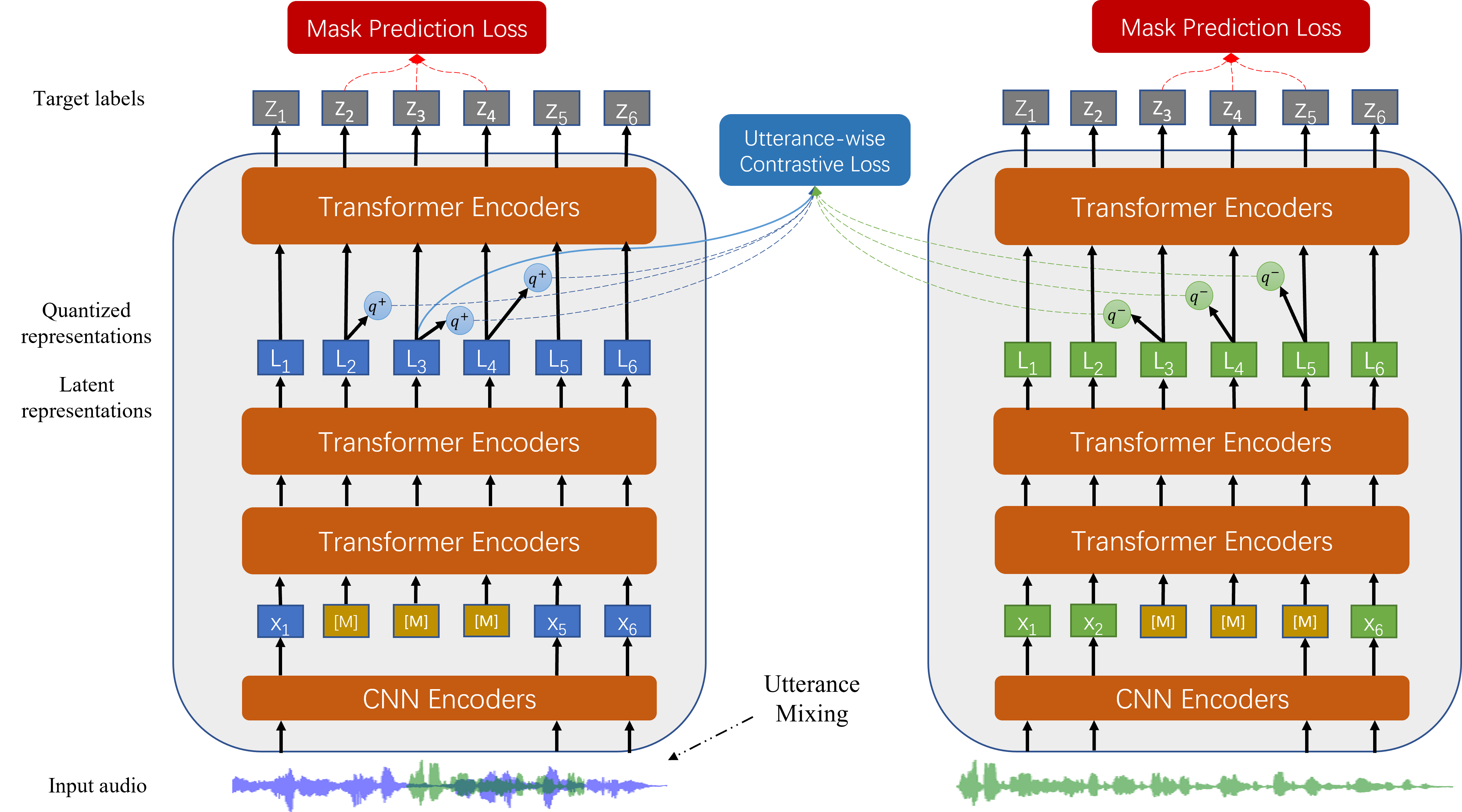}};
	\end{tikzpicture}
	\caption{
	An illustration  of our method. We conduct contrastive loss in the intermediate layer, and use mixed utterance as input. 
		 }\label{fig:UniSpeech-SAT}
\end{figure*}
\vspace{-4mm}

\section{Background}

\label{sec:hubert}

We first overview HuBERT \cite{hsu2021hubert} for universal speech representation learning, which serves as our baseline model. HuBERT has the state-of-the-art performance for several representation learning benchmarks \cite{superb}. 
The main idea of HuBERT is to learn the representation by iterative clustering. HuBERT firstly conducts an offline clustering step based on MFCC (Mel-Frequency Cepstrum Coefficient) of input signal, where the cluster center of each frame is indexed as the pseudo-label for later steps.
Then, a Transformer model with an CNN as a feature extractor is trained on the MFCC and pseudo-labels to form the representation for the first iteration.
A mask prediction loss is used as training criteria, where the network is required to predict the pseudo-label of a masked region from the input sequence, with the features from unmasked parts as input.
Specifically, given a speech utterance with $T$ feature frames, the corresponding labels are $Z = \{z_t\}_{t=1}^T$, the feature sequence $X = \{x_t\}_{t=1}^T$ is extracted from the utterance with CNN Encoders.
We denote $M \subset \{t\}_{t=1}^T$ as the set of masked indices in $\{t\}_{t=1}^T$, $\tilde{X} = r(X, M)$ as the corrupted  $X$ where each $x_t$ is replaced by a random-initialized mask embedding $\tilde{x_t}$ if $t \in M$.
Then the Transformers model $f( \cdot )$ is trained to predict each labels corresponding to the masked indices $\{ z_t | z_t \in Z, t \in M\}$ given the corrupted feature sequence $\tilde{X}$ with the cross-entropy loss $\mathcal{L}_\text{Content} = \sum_{t \in M} \log f(z_t | \tilde{X}, t)$.

The combination of clustering and network training is considered as one iteration. Starting from the second iteration, instead of MFCC feature, the embeddings generated by network from last iteration are used as the input for clustering and network step. 
Presumably both the pseudo-label and the embedding are refined through iterations.

\vspace{-4mm}
\section{UNISPEECH-SAT}
\vspace{-1mm}

We propose \textbf{Uni}versal \textbf{Speech} representation learning with \textbf{S}peaker \textbf{A}ware pre-\textbf{T}raining  (UniSpeech-SAT), which is shown in Figure~\ref{fig:UniSpeech-SAT}.
On top of HuBERT model, two approaches are proposed, namely \textit{the utterance-wise contrastive learning} and \textit{the utterance mixing augmentation}. The former is applied to enhance the single speaker information extraction to improve downstream tasks like speaker verification and speaker identification. The latter mainly benefits the multi-speaker tasks such as speech diarization problem.

\vspace{-4mm}
\subsection{Utterance-wise Contrastive Learning}
\vspace{-1mm}
\label{ssec:utt-loss}

We combine the \textbf{utterance-wise contrastive loss} to enhance unsupervised speaker information modeling.  Two assumptions are made for this integration: 1. Each training utterance contains one active speaker. 2. Each utterance in the training batch belongs to a different speaker, i.e., there is no speaker having two utterances in one batch. 
Given that the dataset is collected from various sources, we believe the two assumptions are mostly satisfied.

In proposed contrastive loss, embeddings within the utterance are considered as positive instances, while the negative instances consists of embedding from other utterances in the same batch.
Suppose that the input feature sequence is $\{ \tilde{X^b} \}_{b=1}^B$, where $B$ is the batch size.
$ \forall {\tilde{X^b}}$, we obtain the latent representation $L^b = \{ l^b_t \}_{t=1}^T$ from the output of an intermediate Transformer encoder layer. 
Then we discretize the latent representation $L^b$ to a finite set of speech representations $Q^b  = \{ q^b_t \}_{t=1}^T$ with a quantization module \cite{wav2vec2}. 
Suppose the quantization module has $G$ codebooks with $V$ entries, we firstly linear transform each latent representations $l$ to logit $l' \in \mathbb{R}^{G \times V}$ and then use Gumbel softmax \cite{jang2016categorical} to choose one discrete entry $e$ from each codebook. The probability for choosing the $v$-th entry from $g$-th codebook is $p_{g,v} = \frac{\exp (l'_{g,v} + n_v) / \tau}{\sum_{k=1}^V \exp (l'_{g,k} + n_k) / \tau} $,
where $\tau$ is  a non-negative temperature, $n_u = - \log (- \log (u))$, and $u$ is uniform sampled from $\mathcal{U}(0, 1)$.
Then we concatenate the selected vectors as $[e_1, \dots, e_G]$, and linear transform it to the quantized representation $q$.
For the latent representation $l^b_t$ centered over mask step $t$ in $b$-th utterance, the model is trained to identify the true quantized representations from the same utterance $\hat{Q^b} = \{ q^b_t | q^b_t \in Q^b, t \in M^b \}$ in a set of quantized candidate representations that uniformly sampled from all the masked time steps in all the utterances within the training batch $\hat{Q} = \cup_{b=1}^B \hat{Q^b}$.
The utterance-wise contrastive loss among $l^b_t$ and $\hat{Q}$ is defined as:
$ \mathcal{L}_\text{Contrastive} =  \sum_{q^b_t \in \hat{Q^b}} \log \frac{\exp(\text{sim}(l^b_t, q^b_t)/\kappa)}{\exp(\text{sim}(l^b_t, q^b_t)/\kappa) + 1}  - \textstyle  \sum_{q^b_t \sim \hat{Q} \setminus \hat{Q^b}} \log \frac{1}{\exp(\text{sim}(l^b_t, q^b_t)/\kappa) + 1}$
, where $\text{sim}(a, b)$ denotes the cosine similarity between the latent representations and quantized representations $a ^\top b / \lVert a \rVert \lVert b \rVert$.
The utterance-wise contrastive loss is augmented by a codebook diversity loss to encourage the equal use of all the codebook entries $\mathcal{L}_\text{d} = \frac{1}{GV} \sum_{g=1}^G \sum_{v=1}^V \bar{p}_{g,v} \log \bar{p}_{g,v}$, 
where $\bar{p}_{g,v}$ is the averaged $p_{g,v}$ across the batch of utterances.
Finally, the speaker information modeling is trained with the loss:
$
 \mathcal{L}_\text{Speaker} = \mathcal{L}_\text{Contrastive} + \alpha \mathcal{L}_\text{d}
$
, where $\alpha$ is a pre-defined hyperparameter. 
Our model will learn the combination of speaker loss and content loss by 
$
 \mathcal{L} = \mathcal{L}_\text{Speaker} + \beta \mathcal{L}_\text{Content} 
$
, where $\beta$ is a pre-defined hyper-parameter.

\vspace{-3mm}
\subsection{Utterance Mixing Augmentation}

\label{ssec:utt-mix}

 We introduce utterance mixing strategy to further boost speaker information modeling in pre-training, especially for multi-speaker tasks such as speaker diarization etc. 
 The utterance mixing method aims to simulate the multi-speaker speech for self-supervised pretraining when only single-speaker pretraining data is available. 
 Specifically, as shown in Algorithm~\ref{algo:utt_mixing}, given a batch of speech utterances $U = \{u^i\}_{i=1}^B$ with batch size $B$, we randomly choose $S$ utterances $\{u^i\}_{i=1}^S$ from the batch. Then for each utterance $u$, we randomly choose an utterance from the batch $u^b \in U$, crop a chunk of random length from $u^b$, and mix it with $u$ in a random region. With the utterance mixing method, the model is trained to extract the information of the main speaker from the mixed audio with the single-speaker information modeling loss (Section~\ref{ssec:utt-loss}), and predict the content information corresponding to the main speaker with the content information modeling loss (Section~\ref{sec:hubert}). 
 Note that we constrain the mixing portion in each utterance to be less than 50\%, avoiding potential label permutation problem.

\begin{algorithm}[t]
\caption{Utterance Mixing}
\footnotesize
\label{algo:utt_mixing}
\begin{algorithmic}[1]
\STATE{\textbf{given} a batch of speech utterances $U = \{u^i\}_{i=1}^B$ with batch size $B$ and length $L$, mixing probability $p$}  
\STATE{Choose $S$ utterances $U^S \subset U$ by Bernoulli sampling with probability $p$}
\FOR{$u$ in $U^S$}
        \STATE{Sample a utterance $u^b$ from discrete uniform distribution with probability $P(u^b = x) = \frac{1}{B}, x \in  U$}
        \STATE{Sample the mix length $l$ from discrete uniform distribution with probability $P(l = x) = \frac{2}{L}, x \in \{1, \cdots, \frac{L}{2}\}$}
        \STATE{Sample a start position $s$ of $u$ from discrete uniform distribution with probability $P(s = x) = \frac{1}{L - l}, x \in \{1, \cdots, L - l\}$}
        \STATE{Sample a start position $s^b$ of $u^b$ from discrete uniform distribution with probability $P(s^b = x) = \frac{1}{L - l}, x \in \{1, \cdots, L - l\}$}
        \STATE{$u[s: s+l] \leftarrow \text{mixing}(u[s: s+l], u^i[s^b: s^b+l])$}
\ENDFOR
\RETURN{$U$}
\end{algorithmic}
\end{algorithm}
\vspace{-4mm}
\subsection{Large and Diverse Pre-training Data}
\label{ssec:94k-data}
We also propose to leverage large-scale unsupervised data from diverse domains to improve the robustness of our model. 
Previous works use Librispeech \cite{librispeech} or Librivox \cite{librilight} datasets for pre-training, which limits the pre-training model since the input data are all extracted from the audiobook. 
We extend the training dataset with (1) 10K hours the Gigaspeech data \cite{GigaSpeech2021}, which is collected from audiobooks, podcasts and YouTube, covering both read and spontaneous speaking styles, and a variety of topics, such as arts, science, sports, etc. (2) 24K hours VoxPopuli data \cite{wang2021voxpopuli}), which from European Parliament (EP) event recordings including plenary sessions, committee meetings and other events. Finally, we have 94k hours data, including LibriVox, VoxPopuli, and Gigaspeech. We believe the diverse dataset can improve model performance on all tasks, because it contains diverse audio background, more speakers, and different contents of speech. 
\begin{table*}[ht]
\vspace{-2mm}
\centering
\caption{
Universal speech representation evaluation on SUPERB benchmark.
The overall score is computed by ourselves: we multiply the QbE score with 100, replace each error rate score with (1 - error rate), and average the scores of all tasks.
\label{table:exp}
}
\vspace{-3mm}
\resizebox{1.0\textwidth}{!}{
\begin{tabular}{l|c|c||c|c|c||c|cc|c|c||c|cc||c||c}

\hline
\multirow{3}{*}{Method} & \multirow{3}{*}{\#Params} & \multirow{3}{*}{Corpus}
 & \multicolumn{3}{c||}{Speaker} & \multicolumn{5}{c||}{Content} & \multicolumn{3}{c||}{Semantics}  & ParaL & Overall \\ \cline{4-16}

& & & SID & ASV & SD & PR & \multicolumn{2}{c|}{ASR (WER)} & KS & QbE & IC & \multicolumn{2}{c||}{SF} & ER & \\ \cline{4-16}

& & & Acc $\uparrow$ & EER $\downarrow$ & DER $\downarrow$ & PER $\downarrow$ & w/o $\downarrow$ & w/ LM $\downarrow$ & Acc $\uparrow$ & MTWV $\uparrow$  & Acc $\uparrow$ & F1 $\uparrow$ & CER $\downarrow$ & Acc $\uparrow$ & Score $\uparrow$ \\ \hline \hline


FBANK & - & - & 8.5E-4 & 9.56 & 10.05 & 82.01 & 23.18 & 15.21 & 8.63 & 0.0058 & 9.10 & 69.64 & 52.94 & 35.39 & 44.2 \\ \hline

PASE+~\cite{pase+} & 7.83M & LS 50 hr & 37.99 & 11.61 & 8.68 & 58.87 & 25.11 & 16.62 & 82.54 & 0.0072 & 29.82 & 62.14 & 60.17 & 57.86 & 57.5  \\ \hline

APC~\cite{apc1} & 4.11M & LS 360 hr & 60.42 & 8.56 & 10.53 & 41.98 & 21.28 & 14.74 & 91.01 & 0.0310 & 74.69 & 70.46 & 50.89 & 59.33 & 67.6  \\

VQ-APC~\cite{vq_apc} & 4.63M & LS 360 hr & 60.15 & 8.72 & 10.45 & 41.08 & 21.20 & 15.21 & 91.11 & 0.0251 & 74.48 & 68.53 & 52.91 & 59.66 & 67.2 \\

NPC~\cite{npc} & 19.38M & LS 360 hr & 55.92 & 9.40 & 9.34 & 20.20 & 13.91 & 43.81 & 88.96 & 0.0246 & 69.44 & 72.79 & 48.44 & 59.08 & 67.0 \\

Mockingjay~\cite{mockingjay} & 85.12M & LS 360 hr & 32.29 & 11.66 & 10.54 & 22.82 & 15.48 & 70.19 & 83.67 & 6.6E-04 & 34.33 & 61.59 & 58.89 & 50.28 & 56.1 \\

TERA~\cite{tera} & 21.33M & LS 360 hr & 57.57 & 15.89 & 9.96 & 18.17 & 12.16 & 49.17 & 89.48 & 0.0013 & 58.42 & 67.50 & 54.17 & 56.27 & 64.2  \\

\hline

modified CPC~\cite{modified_cpc} & 1.84M & LL 60k hr & 39.63 & 12.86 & 10.38 & 42.54 & 20.18 & 13.53 & 91.88 & 0.0326 & 64.09 & 71.19 & 49.91 & 60.96 & 65.1 \\
\hline
wav2vec~\cite{wav2vec} & 32.54M & LS 960 hr & 56.56 & 7.99 & 9.90 & 31.58 & 15.86 & 11.00 & 95.59 & 0.0485 & 84.92 & 76.37 & 43.71 & 59.79 & 71.5 \\

vq-wav2vec~\cite{vq_wav2vec} & 34.15M & LS 960 hr & 38.80 & 10.38 & 9.93 & 33.48 & 17.71 & 12.80 & 93.38 & 0.0410 & 85.68 & 77.68 & 41.54 & 58.24 & 69.3 \\

wav2vec 2.0 Base~\cite{wav2vec2} & 95.04M & LS 960 hr & 75.18 & 5.74 & 6.02 & 6.08 & 6.43 & 4.79 & 96.23 & 0.0233 & 92.35 & 88.30 & 24.77 & 63.43 & 80.3\\

HuBERT Base~\cite{hsu2021hubert} & 94.68M & LS 960 hr & 81.42 & 5.11 & 5.88  & 5.41 & \textbf{6.42} & \textbf{4.79} & 96.30 & 0.0736 & 98.34 & 88.53 & 25.20 & 64.92 & 82.0\\

UniSpeech-SAT Base & 94.68M & LS 960 hr & 85.76 & \textbf{4.31} & 4.41 & 5.40 & 6.75 & 4.86 & 96.75 & 0.0927 & 98.58 & 88.98 & 23.56 & 66.04 & 83.0\\

~~$-$ contrastive loss & 94.68M & LS 960 hr & 84.74 &	4.61&	4.72&	5.22&	6.80 &	5.17&	96.79&	0.0956	&98.31&	88.56	&24.00&	65.60 & 82.8
\\

~~$-$ utterance mixing & 94.68M & LS 960 hr & 85.97	& 4.35& 	5.87& 	5.06	& 7.04& 	5.05& 	96.88& 	0.0866& 	98.10& 	88.50& 	24.52& 	65.97 & 82.7
\\
\hline 

UniSpeech-SAT Base+ & 94.68M & CD 94k hr &  \textbf{87.59}	&  4.36 & 	 \textbf{3.80} & 	\textbf{4.44}	& 6.44 & 	4.88 & 	 \textbf{97.40} &  \textbf{0.1125} & 	 \textbf{98.84} & 	 \textbf{89.76} & 	\textbf{21.75} & 	\textbf{68.48} & \textbf{84.0}

\\
\hline\hline

wav2vec 2.0 Large~\cite{wav2vec2} & 317.38M & LL 60k hr & 86.14 & 5.65 &  5.62 & 4.75 & 3.75 & 3.10 & 96.6 & 0.0489 & 95.28 & 87.11 & 27.31 & 65.64  & 82.1 \\

HuBERT Large~\cite{hsu2021hubert} & 316.61M & LL 60k hr & 90.33 & 5.98 & 5.75 & 3.53 & \textbf{3.62} & \textbf{2.94} & 95.29 & 0.0353 & 98.76 & 89.81 & 21.76 & 67.62 & 83.5 \\

\hline
UniSpeech-SAT Large & 316.61M & CD 94k hr & \textbf{95.16} & \textbf{3.84} &\textbf{3.85} & \textbf{3.38} & 3.99 & 3.19 & \textbf{97.89} & \textbf{0.0836} & \textbf{99.34} & \textbf{92.13} & \textbf{18.01} & \textbf{70.68} & \textbf{85.6} \\

\hline
\end{tabular}
}

\vspace*{-4mm}
\end{table*}
\vspace*{-4mm}
\section{Experiment}
\vspace*{-1mm}
\subsection{Implementation Details}
\vspace*{-1mm}
We implement and pretrain our UniSpeech-SAT model following previous work \cite{hsu2021hubert}.
We pretrain the UniSpeech-SAT Base model for 400k steps on LibriSpeech 960 hours audio \cite{librispeech} using the label generated by clustering the 6-th transformer layer output of the first iteration model of HuBERT Base model.
The UniSpeech-SAT Base+ and UniSpeech-SAT Large model is pretrained for 400k steps on 94K large-scale diverse data (Section~\ref{ssec:94k-data}) using the label generated by clustering the 6-th transformer layer output of the HuBERT Base model.
As for the model architecture and training configurations, we use the same hyperparameters as \cite{hsu2021hubert}. 
\vspace{-5mm}
\subsection{Universal Representation Evaluation}
\vspace{-2mm}
We evaluate our models on SUPERB, which is designed to provide 
a standard and comprehensive testbed for pretrained models on various speech tasks. It covers ten tasks, including Speaker Identification (SID),	 Automatic Speaker Verification (ASV), Speaker Diarization (SD), Phoneme Recognition (PR), Automatic Speech Recognition (ASR), Keyword Spotting (KS), Query by Example Spoken Term Detection (QbE), Intent Classification (IC), Slot Filling (SF), Emotion Recognition (ER). The tasks can be grouped into four aspects
of speech: speaker, content, semantics, and paralinguistics.
We follow the policies created by SUPERB. 1) The design of task specific layers follows the rules of SUPERB. 2) Transformer model is frozen to limit the space of fine-tuning hyper-parameter search. 3) The task specific layer uses the weighted sum results of hidden states from different layers. 

Table \ref{table:exp} shows the evaluation results. There is a significant improvement on speaker diarization task in both base and large setting, where the diarization error rate (DER) is reduced by over 25$\%$. The results demonstrate that the proposed utterance mixing method is very effective for the multi-talker task. Moreover, positive results are observed in speaker identification and speaker verification, which is attributed to the utterance contrastive loss. Surprisingly, our model also obtains substantial gain on emotion recognition. One possible explanation is that the task also requires utterance level information rather than content information. However, our model shows a degradation on ASR without LM. 
The word error rate of our large model is 9$\%$ worse than the baseline, while the gap becomes less than 2$\%$ in the base setting. Our explanation is speaker information and content information orthometric, and the content information is sacrificed given that the model capacity is limited.

\begin{figure}[t]
	\centering
	\includegraphics[width=0.5\textwidth,trim=25 95 73 135,clip]{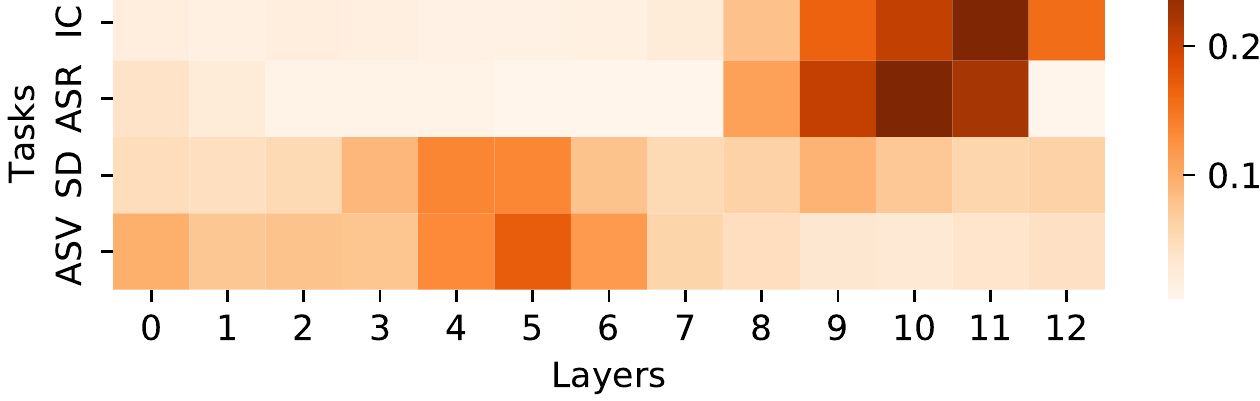}
	\caption{
	Weight Analysis. \label{fig:weight-anaysis}
		 }
\end{figure}
\vspace{-4mm}

\begin{table}[t]
\vspace{-3mm}
\centering
\scriptsize
\setlength{\tabcolsep}{3pt}
\caption{
Results of UniSpeech-SAT Base+ with various mixing ratios on 94k hours training data.  \label{table:exp_ratio}
}
\vspace{-2mm}
\begin{tabular}{c|c|c|cc|c|c}

\hline
\multirow{3}{*}{Method} & \multirow{3}{*}{Ratio} 
 &{Speaker} & \multicolumn{2}{c|}{Content} & \multicolumn{1}{c|}{Semantics}  & ParaL  \\ \cline{3-7}

& & SD & \multicolumn{2}{c|}{ASR (WER)}  & IC & ER \\ \cline{3-7}

& & DER $\downarrow$ & w/o $\downarrow$ & w/ LM $\downarrow$  & Acc $\uparrow$ & Acc $\uparrow$  \\ \hline 

HuBERT Base~\cite{hsu2021hubert} & - & 5.88  & 6.42 & 4.79  & 	 98.34 & 64.92 \\
\hline
\multirow{3}{*}{UniSpeech-SAT Base+} & 0.0 & 	 5.04 & \textbf{6.39} & 	\textbf{4.76}  & 	 99.24 & 	66.32
\\
& 0.2  & 	 3.80	& 6.44 & 	4.88 & 	 98.84 & 	\textbf{68.48} 
\\
& 0.5  & 	 \textbf{3.73} 	& 6.65 & 	5.18  & 	 \textbf{99.29} & 	67.36 
\\
\hline
\end{tabular}
\vspace{-5mm}
\end{table}
\vspace{-2mm}
\subsection{Analysis}
\vspace{-1mm}
\textbf{Weight Analysis: }
Figure \ref{fig:weight-anaysis} shows the layer contribution to different tasks. For speaker verification and diarization, shallow layers contribute more, while for ASR and intent classification, the top layers are more important. The phenomenon indicates the shallow layers learn speaker information while the top layers learn content and semantic information.

\vspace{-0.5mm}

\textbf{Mixing ratio: }
We explore different ratios of mixing utterance and test the performance of mixing 0$\%$, 20$\%$, 50$\%$ utterances, shown in Table \ref{table:exp_ratio}. For 94k hours setting, utterance mixing is still effective. It is a trade-off between speaker and content. We use 20$\%$ for our UniSpeech-SAT Base+ model.

\vspace{-5mm}
\section{Conclusion}
\vspace{-3mm}
In this work, we integrate contrastive loss and utterance mixing to existing framework for unsupervised speech representation learning, aiming at improving the speaker discrimination in learnt embedding. The evaluation on the SUPERB benchmark shows our model achieves the state-of-the-art performance and outperforms other baselines by a large margin.

\bibliographystyle{IEEEbib}
\bibliography{strings,refs}

\end{document}